\documentclass{article}

\usepackage[utf8]{inputenc}
\usepackage[T1]{fontenc}
\usepackage{hyperref}
\usepackage{url}
\usepackage{booktabs}
\usepackage{amsfonts}
\usepackage{amsmath}
\usepackage{amssymb}
\usepackage{graphicx}
\usepackage{xcolor}
\usepackage{microtype}
\usepackage{natbib}
\usepackage[margin=1in]{geometry}

\hypersetup{
    colorlinks=true,
    linkcolor=blue,
    citecolor=blue,
    urlcolor=blue
}

\title{Evolving Deep Learning Optimizers}

\author{
  Mitchell Marfinetz\\
  Independent Researcher\\
  \texttt{mitchmar@sas.upenn.edu}\\
}

\date{}

\begin{document}

\maketitle

\begin{abstract}
We present a genetic algorithm framework for automatically discovering deep learning optimization algorithms.
Our approach encodes optimizers as genomes that specify combinations of primitive update terms (gradient, momentum, RMS normalization, Adam-style adaptive terms, and sign-based updates) along with hyperparameters and scheduling options.
Through evolutionary search over 50 generations with a population of 50 individuals, evaluated across multiple vision tasks, we discover an evolved optimizer that outperforms Adam by 2.6\% in aggregate fitness and achieves a 7.7\% relative improvement on CIFAR-10.
The evolved optimizer combines sign-based gradient terms with adaptive moment estimation, uses lower momentum coefficients than Adam ($\beta_1=0.86$, $\beta_2=0.94$), and notably disables bias correction while enabling learning rate warmup and cosine decay.
Our results demonstrate that evolutionary search can discover competitive optimization algorithms and reveal design principles that differ from hand-crafted optimizers.
Code is available at \url{https://github.com/mmarfinetz/evo-optimizer}.
\end{abstract}

\section{Introduction}
\label{sec:intro}

The choice of optimizer is critical to deep learning success, yet the design of optimization algorithms remains largely a manual process guided by intuition and mathematical analysis.
While optimizers like SGD with momentum \citep{polyak1964some}, Adam \citep{kingma2014adam}, and more recently Lion \citep{chen2023symbolic} have emerged from human insight, the space of possible update rules is vast and largely unexplored.

We propose an evolutionary approach to optimizer discovery that treats the optimization algorithm itself as the object of optimization.
Our key insight is that many successful optimizers can be expressed as weighted combinations of a small set of primitive operations applied to gradients and their running statistics.
By encoding these combinations as genomes and evolving them based on training performance across diverse tasks, we can search for evolved optimizers without requiring manual derivation.

Our contributions are:
\begin{enumerate}
    \item A \textbf{genome representation} for optimization algorithms that captures the essential components of modern optimizers in a compact, evolvable form.
    \item A \textbf{multi-task fitness evaluation} protocol that encourages discovery of optimizers that generalize across datasets and architectures.
    \item \textbf{Empirical evidence} that evolutionary search discovers optimizers competitive with hand-designed algorithms, achieving improvements over Adam on vision benchmarks.
    \item \textbf{Analysis} of the evolved optimizer's structure, revealing design choices (sign-based updates, disabled bias correction, aggressive scheduling) that differ from conventional wisdom.
\end{enumerate}

\section{Related Work}
\label{sec:related}

\paragraph{Hand-Designed Optimizers.}
The progression from SGD \citep{robbins1951stochastic} to momentum \citep{polyak1964some}, AdaGrad \citep{duchi2011adaptive}, RMSProp \citep{tieleman2012lecture}, and Adam \citep{kingma2014adam} represents decades of manual optimizer engineering.
Recent work has produced AdamW \citep{loshchilov2017decoupled}, which decouples weight decay from gradient updates, and Lion \citep{chen2023symbolic}, which uses sign-based updates discovered through symbolic search.

\paragraph{Learning to Optimize.}
\citet{andrychowicz2016learning} introduced the idea of using neural networks to learn optimization algorithms, training an LSTM to output parameter updates.
Subsequent work has explored meta-learning optimizers for specific domains \citep{li2017learning, wichrowska2017learned} and using reinforcement learning to discover update rules \citep{bello2017neural}.
Our approach differs by using genetic algorithms with an explicit, interpretable genome rather than black-box neural networks.

\paragraph{Symbolic and Evolutionary Search.}
\citet{chen2023symbolic} used program search to discover Lion, demonstrating that simple symbolic expressions can outperform complex optimizers.
\citet{real2019regularized} showed evolutionary methods can discover neural architectures competitive with hand-designed ones.
We apply similar evolutionary principles to the optimizer search space.

\section{Method}
\label{sec:method}

\subsection{Genome Representation}

We represent an optimizer as a genome $\mathcal{G}$ that specifies how to compute parameter updates from gradients.
The core update rule is:
\begin{equation}
    \Delta w_t = -\eta_t \sum_{k=1}^{K} \alpha_k \cdot T_k(g_t, m_t, v_t, \epsilon)
\end{equation}
where $\eta_t$ is the (possibly scheduled) learning rate, $\alpha_k$ are learned coefficients, and $T_k$ are primitive terms selected from a catalog.

\paragraph{Primitive Terms.}
We define seven primitive operations that encompass the building blocks of modern optimizers:
\begin{itemize}
    \item \textsc{Grad}: $g_t$ --- raw gradient
    \item \textsc{Momentum}: $m_t$ --- exponential moving average of gradients
    \item \textsc{RmsNorm}: $g_t / (\sqrt{v_t} + \epsilon)$ --- RMSProp-style normalization
    \item \textsc{AdamTerm}: $m_t / (\sqrt{v_t} + \epsilon)$ --- Adam-style adaptive term
    \item \textsc{SignGrad}: $\text{sign}(g_t)$ --- sign of gradient
    \item \textsc{UnitGrad}: $g_t / (|g_t| + \epsilon)$ --- unit gradient
    \item \textsc{Nesterov}: $m_t + \beta_1(g_t - m_t)$ --- Nesterov-style lookahead
\end{itemize}

\paragraph{Genome Components.}
A complete genome consists of:
\begin{itemize}
    \item \textbf{Terms}: List of 1--4 primitive types with corresponding $\alpha$ coefficients
    \item \textbf{Hyperparameters}: $\log_{10}(\eta)$, $\beta_1$, $\beta_2$, $\log_{10}(\epsilon)$, $\log_{10}(\lambda)$ (decoupled weight decay applied in AdamW style, i.e., $w \leftarrow (1 - \eta_t \lambda) w - \Delta w_t$)
    \item \textbf{Flags}: Use momentum ($m_t$), use second moment ($v_t$), bias correction, gradient clipping
    \item \textbf{Schedule}: Warmup steps, cosine decay
\end{itemize}

When gradient clipping is enabled in the genome, we apply elementwise clipping to each gradient component:
$g_t \leftarrow \mathrm{clip}(g_t, -c, c)$ with a fixed threshold $c$.
In the evolved optimizer, clipping is disabled.

This representation can express SGD ($K=1$, \textsc{Grad}), Adam ($K=1$, \textsc{AdamTerm} with bias correction), and novel combinations not previously explored.

\subsection{Fitness Evaluation}

To encourage discovery of general-purpose optimizers, we evaluate each genome across multiple tasks:
\begin{equation}
    \text{Fitness}(\mathcal{G}) = \frac{1}{|T|} \sum_{\tau \in T} \frac{1}{S} \sum_{s=1}^{S} f(\mathcal{G}, \tau, s)
\end{equation}
where $T$ is the set of tasks, $S$ is the number of random seeds, and $f(\mathcal{G}, \tau, s)$ is the scalar fitness on task $\tau$ with seed $s$.
Concretely, we define
\begin{equation}
    f(\mathcal{G}, \tau, s) =
    \begin{cases}
        \mathrm{acc}_{\text{test}} + \lambda_{\text{loss}} \cdot \max\bigl(0, 1 - \overline{\ell}_{\text{train,last 50}}\bigr) & \text{if run is stable},\\
        -1 & \text{if run diverges},
    \end{cases}
    \label{eq:fitness_term}
\end{equation}
where $\mathrm{acc}_{\text{test}}$ is test accuracy at the final step, $\overline{\ell}_{\text{train,last 50}}$ is the mean training loss over the last 50 steps, and $\lambda_{\text{loss}} = 0.05$.
This bonus is capped at $+0.05$ and is only positive when the final training loss is below $1.0$, which explains why fitness values can slightly exceed $1.0$ on MNIST.
We mark a run as divergent if the training loss becomes NaN or Inf, exceeds $50$ at any point, or if the training loop raises a numerical \texttt{RuntimeError}; such runs receive fitness $-1$.

\subsection{Genetic Algorithm}

We evolve a population of $N=50$ genomes over $G=50$ generations using:

\paragraph{Selection.}
Tournament selection with $k=4$ candidates, plus elitism preserving the top 3 individuals.

\paragraph{Crossover.}
Single-point crossover for term lists; uniform crossover for scalar hyperparameters and flags.

\paragraph{Mutation.}
Adaptive mutation rates that decay over generations:
\begin{itemize}
    \item Numeric mutations: Gaussian perturbation of hyperparameters
    \item Structural mutations: Add, remove, or change primitive terms
    \item Flag mutations: Flip boolean settings
\end{itemize}

\paragraph{Initialization.}
The initial population includes known optimizers (SGD, Adam, AdamW, RMSProp) as seeds, with the remainder randomly generated.

\section{Experiments}
\label{sec:experiments}

\subsection{Setup}

\paragraph{Tasks.}
We evaluate on three vision classification tasks:
\begin{itemize}
    \item \textbf{Fashion-MNIST}: 28$\times$28 grayscale, 10 classes, SmallCNN
    \item \textbf{CIFAR-10}: 32$\times$32 RGB, 10 classes, deeper CNN with BatchNorm
    \item \textbf{MNIST}: 28$\times$28 grayscale, 10 classes, SmallCNN
\end{itemize}

\paragraph{Training.}
During evolution, each evaluation trains for 500 optimization steps with batch size 128.
For the final comparison in Table~\ref{tab:results}, we re-train each optimizer for 1000 steps using the same architectures, batch size, and data subsampling protocol.
Evolution uses 2 seeds per task, while the final comparison averages over 3 seeds.
Total evolution time was approximately 13 hours on an NVIDIA Tesla T4.

For each dataset we sample a fixed random subset of 15{,}000 training examples to speed up evaluation.
All optimizers, including baselines, are trained on the same subset for a given task.

\paragraph{Architectures and Implementation Details.}
For MNIST and Fashion-MNIST we use a SmallCNN with two convolutional layers (32 and 64 channels, $3\times 3$ kernels, ReLU activations), each followed by $2\times 2$ max-pooling, and a classifier consisting of a fully connected layer with 256 hidden units, ReLU, dropout, and a final linear layer to 10 classes.
For CIFAR-10 we use a CNN with two convolutional blocks: a 32-channel block and a 64-channel block, each containing two $3\times 3$ convolutions with Batch Normalization and ReLU, followed by $2\times 2$ max-pooling and dropout, and a classifier with a 512-unit fully connected layer, dropout, and a final linear classifier.
All models use cross-entropy loss.
We implement optimizers in PyTorch and apply decoupled weight decay and (optionally) gradient clipping as described in Sec.~\ref{sec:method}.

\subsection{Results}

\begin{table}[t]
\centering
\caption{Comparison of evolved optimizer with baselines. Fitness combines test accuracy with a small training-loss bonus (Eq.~\ref{eq:fitness_term}). Results averaged over 3 seeds with 1000 training steps.}
\label{tab:results}
\begin{tabular}{lcccc}
\toprule
\textbf{Optimizer} & \textbf{Fashion-MNIST} & \textbf{CIFAR-10} & \textbf{MNIST} & \textbf{Overall} \\
\midrule
SGD (momentum) & 0.826 & 0.467 & 0.988 & 0.760 \\
RMSProp & 0.928 & 0.546 & 1.035 & 0.836 \\
AdamW & 0.934 & 0.640 & 1.035 & 0.870 \\
Adam & 0.935 & 0.647 & 1.036 & 0.872 \\
\midrule
\textbf{Evolved} & \textbf{0.949} & \textbf{0.697} & \textbf{1.039} & \textbf{0.895} \\
\bottomrule
\end{tabular}
\end{table}

Table~\ref{tab:results} shows that the evolved optimizer outperforms all baselines, with the largest improvement on CIFAR-10 (+7.7\% relative to Adam).

\subsection{Analysis of Evolved Optimizer}

The best genome after 50 generations has the following structure:

\paragraph{Update Rule.}
\begin{equation}
    \Delta w = -\eta \left(
      0.73\,\mathrm{sign}(g)
      + 3.63\,\frac{m}{\sqrt{v} + \epsilon}
    \right)
\end{equation}

This combines sign-based updates (total weight 0.73) with Adam-style adaptive terms (total weight 3.63).
The underlying genome contained duplicate primitive terms (two SignGrad and two AdamTerm entries); we merge them here for clarity by summing their coefficients.

\paragraph{Hyperparameters.}
\begin{itemize}
    \item Learning rate: $1.2 \times 10^{-3}$ (slightly higher than Adam's typical $10^{-3}$)
    \item $\beta_1 = 0.855$ (vs. Adam's 0.9) --- less momentum smoothing
    \item $\beta_2 = 0.936$ (vs. Adam's 0.999) --- faster adaptation to gradient magnitude
    \item Weight decay: $9.7 \times 10^{-4}$
\end{itemize}

\paragraph{Notable Design Choices.}
\begin{itemize}
    \item \textbf{Bias correction disabled}: Unlike Adam, the evolved optimizer does not correct for initialization bias in moment estimates.
    \item \textbf{Warmup enabled}: 100 steps of linear warmup, which may compensate for disabled bias correction.
    \item \textbf{Cosine decay}: Learning rate annealing over training.
    \item \textbf{Sign gradient component}: Provides magnitude-invariant updates, similar to Lion.
\end{itemize}

\begin{figure}[t]
    \centering
    \includegraphics[width=0.8\textwidth]{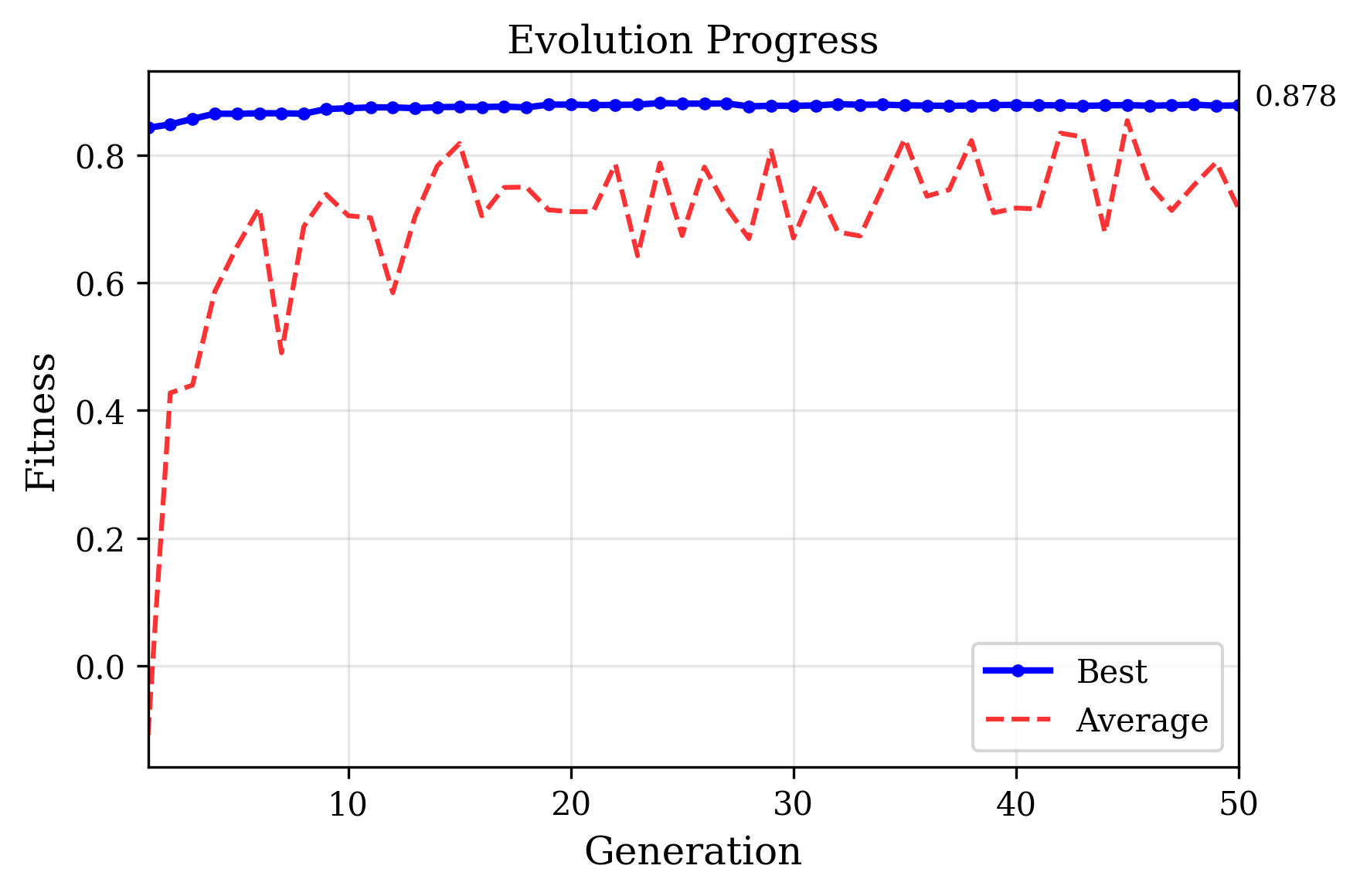}
    \caption{Evolution progress showing best and average fitness over 50 generations.}
    \label{fig:evolution}
\end{figure}

\paragraph{Population Dynamics.}
Figure~\ref{fig:evolution} shows that \textsc{SignGrad} and \textsc{AdamTerm} dominated the population by generation 50, with 102 and 88 occurrences respectively.
Other primitives were largely eliminated by selection pressure, suggesting these two components are particularly effective.

\section{Discussion}
\label{sec:discussion}

\paragraph{Relation to Lion.}
The evolved optimizer's use of sign-based updates parallels the Lion optimizer \citep{chen2023symbolic}, which was discovered through symbolic program search.
However, our approach discovered this independently through evolutionary pressure, providing convergent evidence for the effectiveness of sign-based updates.
Unlike Lion, our evolved optimizer retains Adam-style adaptive terms, suggesting a hybrid approach may be beneficial.

\paragraph{Why Disable Bias Correction?}
Adam's bias correction compensates for the zero-initialization of moment estimates.
Our evolved optimizer disables this but enables warmup, which serves a similar purpose by using small learning rates early in training.
This suggests the two mechanisms may be redundant.

\paragraph{Lower Momentum Coefficients.}
The evolved $\beta_1 = 0.855$ and $\beta_2 = 0.936$ make the optimizer more responsive to recent gradients than Adam.
This may be beneficial for the relatively short training runs (500 steps) used during evolution, though it warrants investigation on longer training.

\section{Limitations and Future Work}
\label{sec:limitations}

\begin{itemize}
    \item \textbf{Scale}: We evaluated on small CNNs with short training runs. Validation on larger models (ResNets, Transformers) and longer training is needed.
    \item \textbf{Task diversity}: Only vision classification tasks were used. Language modeling and reinforcement learning would test generalization.
    \item \textbf{Single evolution run}: Results are from one evolutionary run. Multiple runs would establish robustness.
    \item \textbf{Missing baselines}: We did not compare to Lion, AdaFactor, or other modern optimizers.
    \item \textbf{Theoretical analysis}: We provide no convergence guarantees or theoretical justification for the evolved update rule.
\end{itemize}

Future work could address these limitations and explore whether the genome representation can be extended to capture more complex scheduling strategies or per-layer adaptation.

\section{Conclusion}
\label{sec:conclusion}

We presented a genetic algorithm framework for discovering deep learning optimizers.
By encoding optimizers as genomes specifying combinations of primitive update terms, we evolved an optimizer that outperforms Adam on vision benchmarks.
The evolved algorithm combines sign-based and adaptive updates, uses aggressive scheduling, and makes design choices (disabled bias correction, lower momentum) that differ from hand-designed optimizers.
Our results suggest that evolutionary search is a viable approach to optimizer discovery and can reveal design principles not apparent from manual analysis.


\bibliographystyle{plainnat}
\bibliography{references}

\begin{thebibliography}{12}
\providecommand{\natexlab}[1]{#1}
\providecommand{\url}[1]{\texttt{#1}}
\expandafter\ifx\csname urlstyle\endcsname\relax
  \providecommand{\doi}[1]{doi: #1}\else
  \providecommand{\doi}{doi: \begingroup \urlstyle{rm}\Url}\fi

\bibitem[Andrychowicz et~al.(2016)Andrychowicz, Denil, Gomez, Hoffman, Pfau,
  Schaul, Shillingford, and De~Freitas]{andrychowicz2016learning}
Marcin Andrychowicz, Misha Denil, Sergio Gomez, Matthew~W Hoffman, David Pfau,
  Tom Schaul, Brendan Shillingford, and Nando De~Freitas.
\newblock Learning to learn by gradient descent by gradient descent.
\newblock In \emph{Advances in Neural Information Processing Systems}, pages
  3981--3989, 2016.

\bibitem[Bello et~al.(2017)Bello, Zoph, Vasudevan, and Le]{bello2017neural}
Irwan Bello, Barret Zoph, Vijay Vasudevan, and Quoc~V Le.
\newblock Neural optimizer search with reinforcement learning.
\newblock In \emph{International Conference on Machine Learning}, pages
  459--468, 2017.

\bibitem[Chen et~al.(2023)Chen, Liang, Huang, Real, Wang, Liu, Pham, Dong,
  Luong, Hsieh, Lu, and Le]{chen2023symbolic}
Xiangning Chen, Chen Liang, Da~Huang, Esteban Real, Kaiyuan Wang, Yao Liu, Hieu
  Pham, Xuanyi Dong, Thang Luong, Cho-Jui Hsieh, Yifeng Lu, and Quoc~V Le.
\newblock Symbolic discovery of optimization algorithms.
\newblock \emph{arXiv preprint arXiv:2302.06675}, 2023.

\bibitem[Duchi et~al.(2011)Duchi, Hazan, and Singer]{duchi2011adaptive}
John Duchi, Elad Hazan, and Yoram Singer.
\newblock Adaptive subgradient methods for online learning and stochastic
  optimization.
\newblock \emph{Journal of Machine Learning Research}, 12:\penalty0 2121--2159,
  2011.

\bibitem[Kingma and Ba(2014)]{kingma2014adam}
Diederik~P Kingma and Jimmy Ba.
\newblock Adam: A method for stochastic optimization.
\newblock \emph{arXiv preprint arXiv:1412.6980}, 2014.

\bibitem[Li and Malik(2017)]{li2017learning}
Ke~Li and Jitendra Malik.
\newblock Learning to optimize.
\newblock In \emph{International Conference on Learning Representations}, 2017.

\bibitem[Loshchilov and Hutter(2017)]{loshchilov2017decoupled}
Ilya Loshchilov and Frank Hutter.
\newblock Decoupled weight decay regularization.
\newblock \emph{arXiv preprint arXiv:1711.05101}, 2017.

\bibitem[Polyak(1964)]{polyak1964some}
Boris~T Polyak.
\newblock Some methods of speeding up the convergence of iteration methods.
\newblock \emph{USSR Computational Mathematics and Mathematical Physics},
  4\penalty0 (5):\penalty0 1--17, 1964.

\bibitem[Real et~al.(2019)Real, Aggarwal, Huang, and Le]{real2019regularized}
Esteban Real, Alok Aggarwal, Yanping Huang, and Quoc~V Le.
\newblock Regularized evolution for image classifier architecture search.
\newblock In \emph{AAAI Conference on Artificial Intelligence}, volume~33,
  pages 4780--4789, 2019.

\bibitem[Robbins and Monro(1951)]{robbins1951stochastic}
Herbert Robbins and Sutton Monro.
\newblock A stochastic approximation method.
\newblock \emph{The Annals of Mathematical Statistics}, pages 400--407, 1951.

\bibitem[Tieleman and Hinton(2012)]{tieleman2012lecture}
Tijmen Tieleman and Geoffrey Hinton.
\newblock Lecture 6.5-rmsprop: Divide the gradient by a running average of its
  recent magnitude.
\newblock COURSERA: Neural Networks for Machine Learning, 2012.

\bibitem[Wichrowska et~al.(2017)Wichrowska, Maheswaranathan, Hoffman,
  Colmenarejo, Denil, de~Freitas, and Sohl-Dickstein]{wichrowska2017learned}
Olga Wichrowska, Niru Maheswaranathan, Matthew~W Hoffman, Sergio~Gomez
  Colmenarejo, Misha Denil, Nando de~Freitas, and Jascha Sohl-Dickstein.
\newblock Learned optimizers that scale and generalize.
\newblock In \emph{International Conference on Machine Learning}, pages
  3751--3760, 2017.

\end{thebibliography}

\end{document}